%% file: tpp-wsdm.tex
\documentclass[sigconf]{acmart}

\RequirePackage{natbib}
\usepackage{xcolor}
\usepackage{enumitem}
\usepackage{graphicx,dsfont,bm,subfig}
\RequirePackage[skip=.8ex plus.2ex]{caption}
\RequirePackage{tikz}
\RequirePackage{pgfplots}
\pgfplotsset{compat=newest}
\usepgfplotslibrary{groupplots}
\usepgfplotslibrary{fillbetween}
\newcommand{\nn}{\nonumber}
\RequirePackage[color=yellow!10,textsize=scriptsize,linecolor=gray!80]{todonotes} 
\usepackage[english]{babel}
\usepackage{latexsym}
\usepackage{mathrsfs}
\usepackage{bbm}
\usepackage{etoolbox}
\usepackage{cleveref}
\usepackage[vlined,ruled]{algorithm2e}
\usepackage{algorithmic}
\usepackage{url}
\usepackage{adjustbox}
\usepackage{multirow}

\RequirePackage{Definitions}

\graphicspath{{./FIG/}}
\def\our{{\scshape DualTPP}\xspace}

\newcommand{\RmtppMseOpt}{\our}
\newcommand{\RmtppMsevarOpt}{\our}

\newcommand{\RmtppNllSimu}{RMTPP \xspace}
\newcommand{\RmtppMseOptComp}{Hierarchical \xspace}
\newcommand{\RmtppMsevarOptComp}{Hierarchical \xspace}
\newcommand{\WGAN}{WGAN \xspace}
\newcommand{\SeqSeq}{Seq2Seq \xspace}
\newcommand{\Transformer}{TransMTPP\xspace}

\newcommand{\CountOnly}{Count-only\xspace}
\newcommand{\MuOnlyOpt}{\our-without-count-variance\xspace}
\newcommand{\RmtppNllOpt}{\our-with-intensity\xspace}
\newcommand{\RmtppMseSimu}{Event-only \xspace}
\newcommand{\RmtppMsevarSimu}{Event-only \xspace}

\copyrightyear{2021} 
\acmYear{2021} 
\setcopyright{acmcopyright}\acmConference[WSDM '21]{Proceedings of the Fourteenth ACM International Conference on Web Search and Data Mining}{March 8--12, 2021}{Virtual Event, Israel}
\acmBooktitle{Proceedings of the Fourteenth ACM International Conference on Web Search and Data Mining (WSDM '21), March 8--12, 2021, Virtual Event, Israel}
\acmPrice{15.00}
\acmDOI{10.1145/3437963.3441740}
\acmISBN{978-1-4503-8297-7/21/03}

\begin{document}

\title{Long Horizon Forecasting With Temporal Point Processes}
 
\author{Prathamesh Deshpande}
\affiliation{
  \institution{IIT Bombay}}
\authornote{Contact author, pratham@cse.iitb.ac.in}
 
\author{Kamlesh Marathe}
\affiliation{
  \institution{IIT Bombay}}

\author{Abir De}
\affiliation{
  \institution{IIT Bombay}}

\author{Sunita Sarawagi}
\affiliation{
  \institution{IIT Bombay}}

\begin{abstract}
In recent years, marked temporal point processes (MTPPs) have emerged as a powerful modeling machinery to  
characterize asynchronous events in a wide variety of applications. 
MTPPs have demonstrated significant potential in predicting event-timings, especially for events arriving in near future.  
However, due to current design choices, MTPPs often show poor predictive performance at forecasting event arrivals in distant future. 
To ameliorate this limitation, in this paper, we design \our\ which is specifically well-suited to long horizon event forecasting. \our\ has two components. The first component is an intensity free MTPP model, which captures microscopic 
event dynamics by modeling the time of future events. The second component takes a different dual perspective of modeling aggregated counts of events in a given time-window, thus encapsulating macroscopic event dynamics. Then we develop a novel inference framework jointly over the two models 
by solving a sequence of constrained quadratic optimization problems. Experiments with a diverse set of real datasets show that \our\ outperforms existing MTPP methods on long horizon forecasting by substantial margins, achieving almost an order of magnitude reduction in Wasserstein distance between actual events and forecasts.
The code and the datasets can be found at the following URL: \url{https://github.com/pratham16cse/DualTPP}

\end{abstract}

\maketitle


\section{Introduction}
\label{sec:introduction}
In recent years, marked temporal point processes (MTPPs) have emerged as a powerful tool in modeling asynchronous events  
in a diverse set of applications, such as information diffusion
in social networks~\cite{,kong2020modeling,lamprier2019recurrent,de2018demarcating,rizoiu2017online,zhou2013learning,du2013uncover,yang2013mixture,NIPS2012_4582},
disease progression~\cite{qian2020learning,yang2019modeling,rizoiu2018sir,saichev2011generating},
traffic flow~\cite{okawa2019deep}, and
financial transactions~\cite{hawkes2020hawkes,hambly2019spde,maciak2019infinitely,giesecke2018filtered,trinh2018non,filimonov2015apparent}.
MTPPs are realized using two quantities: (i) intensity functions which characterize the probabilities of arrivals of subsequent events, based on the history of previous events; and   
(ii) the distribution of marks which captures extra information attached with each  event \eg, sentiment in a Tweet, location in traffic flow, etc.
%
%
%

Over the myriad applications of MTPPs, we identify two modes in which MTPPs are used during prediction:
(i) nowcasting, which implies
prediction of only the immediate next event
i.e.\ one-step ahead prediction
; and, (ii) forecasting, which requires prediction of events in a distant future
i.e.\ long-term forecasting
.
%
%
Forecasting continuous-time events with TPP models has 
a wide variety of use cases. For example,
in emergency planning, it can assist resource allocation by anticipating demand; in transportation, it can help in congestion management; and
in a social network, it can help to anticipate the rise of
an orchestrated campaign. 
In this work, our goal is to develop a temporal point process model, which is specifically suited for accurate forecasting
of arrival of events in the long term given a history of events in the past.

\xhdr{Limitations of prior work}
Predictive models of temporal point processes have been extensively researched in recent literature~\cite{mei2017neural,du2016recurrent,omi2019fully,loaiza2019deep,xiao2019learning,apostolopoulou2019mutually,upadhyay2018deep,cai2018modeling,xiao2017joint,vassoy2019time,zhong2018time}. 
%
A predominate approach is to train an intensity function for the next event conditioned on historical events, and then based on this estimated intensity function, forward sample events to predict a sequence of events in the future.
While  these approaches have 
shown promise at predicting the arrival
of 
events in the near future, they suffer from two major limitations:
\begin{itemize}[nolistsep, noitemsep, leftmargin=*]
 \item[I]  Their modeling frameworks heavily rest on designing the intensity function--- which in turn can sample only the next subsequent event.
Such a design choice allows these models to be trained only for nowcasting rather than forecasting.
 \item[II]  Over long time horizons, the forward sampling method accumulates cascading errors as we condition on predicted events to generate the next event, whereas during training we condition on true events.  Existing approaches~\cite{xiao2017wasserstein, xiao2018learning} of handling this mismatch via sequence-level losses provide only modest gains.
 
\end{itemize}
%
%

\xhdr{Present work} 
Responding to the limitations of prior approaches, we develop \our, which is specifically designed to forecast events over long time horizons. %
The \our\ model consists of two components.  The first component encapsulates the event dynamics at a microscopic scale, whereas the second component views the same event dynamics 
from a different perspective and at a higher macroscopic scale.
The first component is an intensity free recurrent temporal point process, which models the \emph{time} of events conditioned on all previous events along with marks.  This model has sufficient predictive ability to capture the event arrival process in the immediate future, but like existing TPPs is subject to cascading drift.
The second component models the \emph{count} of events over fixed time-intervals in the long-term future.  
Together, this leads to  an accurate modeling of both short and long term behavior of the associated event arrival process.  
%
%
%
%

Inference in \our involves forecasting events while achieving consensus across predictions from both models.  This 
presents new algorithmic challenges. We formulate a novel joint inference objective on the two models, and show how to decompose it into a sequence of constrained concave quadratic maximization problems over continuous variables, combined with a binary search over discrete count variables.
Our algorithm provides a significant departure from existing sampling-based inference that are subject to gross inaccuracies.

Our model includes both elements of multi-scale modeling like in hierarchies and  multi-view learning.
%
We show that this form of multi-view, multi-scale modeling, coupled with our joint inference algorithm, provides more accurate long-term forecasting than just multi-scale models~\cite{wavenets,borovykh2017conditional,vassoy2019time}.
%
%
%
We provide a comprehensive evaluation of our proposal across several real world datasets.
Our experiments show that the proposed model outperforms several state-of-the-art baselines in terms of forecasting accuracy, by a substantial margin.

\xhdr{Summary of contributions} 
Summarizing, we make the following contributions in this paper.\\
\emph{--- Forecasting aware modeling framework:} We propose a novel
forecasting aware modeling framework for temporal point process, which
consists of two parts--- the first part captures the low-level microscopic behavior, whereas the other part captures the high level macroscopic signals from a different perspective. These two components complement the predictive ability of each other, that helps the joint model to accurately characterize the 
long 
horizon behavior of the event dynamics.\\
\emph{--- Efficient inference protocol:} We devise a novel inference method to forecast the arrival of 
events during an arbitrary time-interval. In sharp contrast to expensive sampling procedures, the proposed inference method casts the forecasting task as a 
sequence of 
constrained quadratic optimization problems, which can be efficiently solved using standard tools.  \\
\emph{--- Comprehensive evaluation: }Our proposal is not only theoretically principled, but also practically effective. We show superior predictive ability compared to several state-of-the-art algorithms. Our experiments are on practically motivated datasets spanning applications in social media, traffic and emergency planning.  The substantial gains we obtain over existing methods establish our practical impact on these applications.
\section{Related work}
Our work is related to temporal point processes, long-term forecasting in time-series, and peripherally with the area of multi-view learning.

\xhdr{Temporal point process} 
Modeling continuous time event streams with temporal point processes (TPPs) follow two predominant approaches.
The first approach focuses on characterizing TPPs using fixed parameterizations,
by means of linear or quasi-linear forms of intensity functions~\cite{hawkes1971spectra,hawkes1971point,ogata1998space,bernardo2003markov,isham1979self}, \eg, Hawkes process, self-correcting process, etc.
Such TPP models are designed  to capture specific phenomena of interest. For example, Hawkes process 
encapsulates the self-exciting nature of information diffusion in online social networks whereas, Markov modulated point process 
can accurately model online check-ins. 
While such models provide interpretability, their fixed parameterizations
often lead to model mis-specifications, limited expressiveness, which in turn constrain their predictive power. The second approach 
overcomes such limitations by designing deep neural TPP models, guided by a recurrent neural network which captures the dependence of
previous events 
on 
the arrival of subsequent events. \citet{du2016recurrent} proposed Recurrent Marked Temporal Point Process (RMTPP), a three layer neural architecture for TPP model, which relies on a vanilla RNN to capture the dependence between 
inter-event arrival times.
Such a design is still the workhorse of many deep recurrent TPP models. Neural Hawkes process~\cite{mei2017neural}
provides a robust nonlinear TPP model, which can incorporate the effect of missing data. 
However, these models heavily rest on learning the arrival dynamics of one subsequent event and as a consequence,
they show poor forecasting performance.
Recently, a number of more powerful deep learning techniques have been borrowed to capture richer dependencies among events in TPPs.
For example,~\citet{xiao2018learning} proposes a sequence to sequence encoder-decoder model for predicting \emph{next $k$} events;
\citet{xiao2017wasserstein} use Wasserstein GANs to generate an entire sequence of events;
\citet{vassoy2019time} deploy a hierarchical model; and
~\citet{zuo2020transformer} apply transformer architecture to capture the dependence among events via self-attention.  We compare \our against these methods in Section~\ref{sec:expts} and show substantial gains. 

\xhdr{Long-term Forecasting in Time Series}
The topic of long-term forecasting has been more explored in the regular time-series setting than in the TPP setting.  Existing time-series models are also auto-regressive and trained for one-step ahead prediction~\cite{FlunkertSG17}, and subject to similar phenomenon of cascading errors when used for long-range forecasting. Efforts to fix the teacher-forcing training of these one-step ahead model to adapt better to multi-step forecasting~\cite{Venkatraman15}, have been not as effective as breaking the auto-regressive structure to
directly predict for each future time-step~\cite{Taieb2015,wen2017multi, Deshpande2019}. Another idea is to use dilated convolutions, as successfully deployed in Wavenet~\cite{wavenets} for audio generation, that connect each output to successively doubling hops into the past~\cite{borovykh2017conditional}. A hierarchical model that we compared with in Section~\ref{sec:expt:baselines} also uses dilated connections to past events. We found that this model provided much better long-range forecasts than existing TPP models, however our hybrid event-count model surpassed it consistently.
A third idea is to use a loss function~\cite{Guen2019} over the entire prediction range that preserves sequence-level properties, analogous to how Wasserstein loss is used in \cite{xiao2018learning} for the TPP setting.

A key difference of \our\ compared to all previous work in both the TPP and time-series literature is that, all existing methods focus on training, and during inference continue to deploy the same one-step event generation. Our key idea is to use a second model to output properties of the aggregated set of predicted events. We then solve an efficient joint optimization problem over the predicted sequence to achieve consensus between the predicted aggregate properties and one-step generated events. This relates our approach to early work on multi-view learning in the traditional machine learning literature that we discuss next.

\xhdr{Multi-view Learning Models}
Inference in structured prediction tasks with aggregate potentials over a large number of predicted variables was studied in tasks like image segmentation, \cite{RamalingamKAT08,KohliLT09,Tarlow10} and information extraction~\cite{gupta10}. In several NLP tasks too, enforcing constraints during inference via efficient optimization formulations has been found to be effective in \cite{PRYZ05Learning,
FersiniMFR14,DeutschUR19}. In this paper we demonstrate, for the first time, the use of these ideas to TPPs, which due to their continuous nature, pose very different challenges than classical multi-view models on discrete labels. 
 
\section{Model formulation}
In this section, we formulate \our, our two-component modeling framework for marked temporal point processes (MTPPs). 
We begin with an overview of MTPPs and then provide a detailed description of our proposed \our. 
\subsection{Background on MTPP}
\label{sec:overviewMTPP}
An MTPP~\cite{du2016recurrent,xiao2017wasserstein,zuo2020transformer}  is a  stochastic process, which is realized using a series of discrete events arriving in continuous time. 
Given a sequence of events $\set{e_1=(m_1,t_1), e_2=(m_2,t_2), \ldots}$ where  $m_i\in [K]$ 
\footnote{\scriptsize In the current work, we consider discrete marks which can take $K$ labels, however, our method can easily be extended to continuous marks.} indicate the discrete mark  and $t_i\in\RR^+$ indicate the arrival time of the $i-$th event, an MTPP is characterized by
$\Hcal_t=\set{e_i=(m_i,t_i)|t_i<t}$ which gathers all events that arrived until time $t$. Equivalently, it can also be  described using a counting process $N(t)$ which counts the number of events arrived until time $t$, \ie, $N(t)=|\Hcal_t|$. 
The dynamics of $N(t)$ is characterized using an intensity function
$\lambda^*(t)$, which specifies the likelihood of the next event, conditioned on the history of events $\Hcal_t$\footnote{\scriptsize $^*$ indicates the dependence on history}.  
%
The intensity function
$\lambda^*(t)$ computes the infinitesimal probability that an event will happen in the time window $(t,t+dt]$ conditioned on the history $\Hcal_t$ as follows:
\begin{align}
    \PP(dN(t)=N(t+dt)-N(t)=1\,|\, \Hcal_t) = \lambda^*(t) dt,
\end{align}
The intensity function is used to compute the expected time of the next event as:
\begin{align}
    \EE[t_i\,|\, \Hcal_{t_i}] = \int_{t_{i-1}} ^{\infty} t\cdot \lambda^*(t) dt
\end{align}
The marks are generated using some probability distribution $q_m$ conditioned on the history of the events, \ie, 
\begin{align}
\PP(m_i=k|\Hcal_{t_i})=q^* _m(k) 
\end{align}
Given the history of events $\Hcal_T$ observed during the time interval $(0,T]$, one typically learns the intensity function $\lambda^*(t)$ and the mark distribution $q_m ^*$ by maximizing the following likelihood function:
\begin{align}
    \Lcal(\Hcal_T\,|\, \lambda^*, q_m ^*) = \sum_{(m_i,t_i)\in\Hcal_T}\bigl(\log q_m ^* (m_i)+\log \lambda^* (t_i)\bigr) + \int_0 ^T \lambda^*(\tau) d\tau \nn
\end{align}

%
Once the intensity function $\lambda^*(t)$ and the mark distribution $q^*_{m}$ are estimated, 
they are used to forecast events by means of thinning~\cite{ogata1998space} or 
inverse sampling~\cite{upadhyay2018deep} mechanisms. 
Such mechanisms often suffer from  poor time complexity. Moreover, such recursive sampling methods build up prediction error led by any model mis-specification.
In the following, we aim to design a temporal point process model that is able to overcome that limitation.

\subsection{Design of \our}
\label{subsec:model}
We now set about to design our proposed model \our. 
At the very outset, \our\ has two components to model the underlying MTPP--- the event model for capturing the dynamics of individual events, 
and the count model that provides an alternative count perspective over a set of events in the long-term future.  Here we describe the model structure and training. In Section~\ref{sec:infer}, we describe how we combine outputs from the two models during inference.
%
\begin{figure*}
\includegraphics[width=0.65\textwidth]{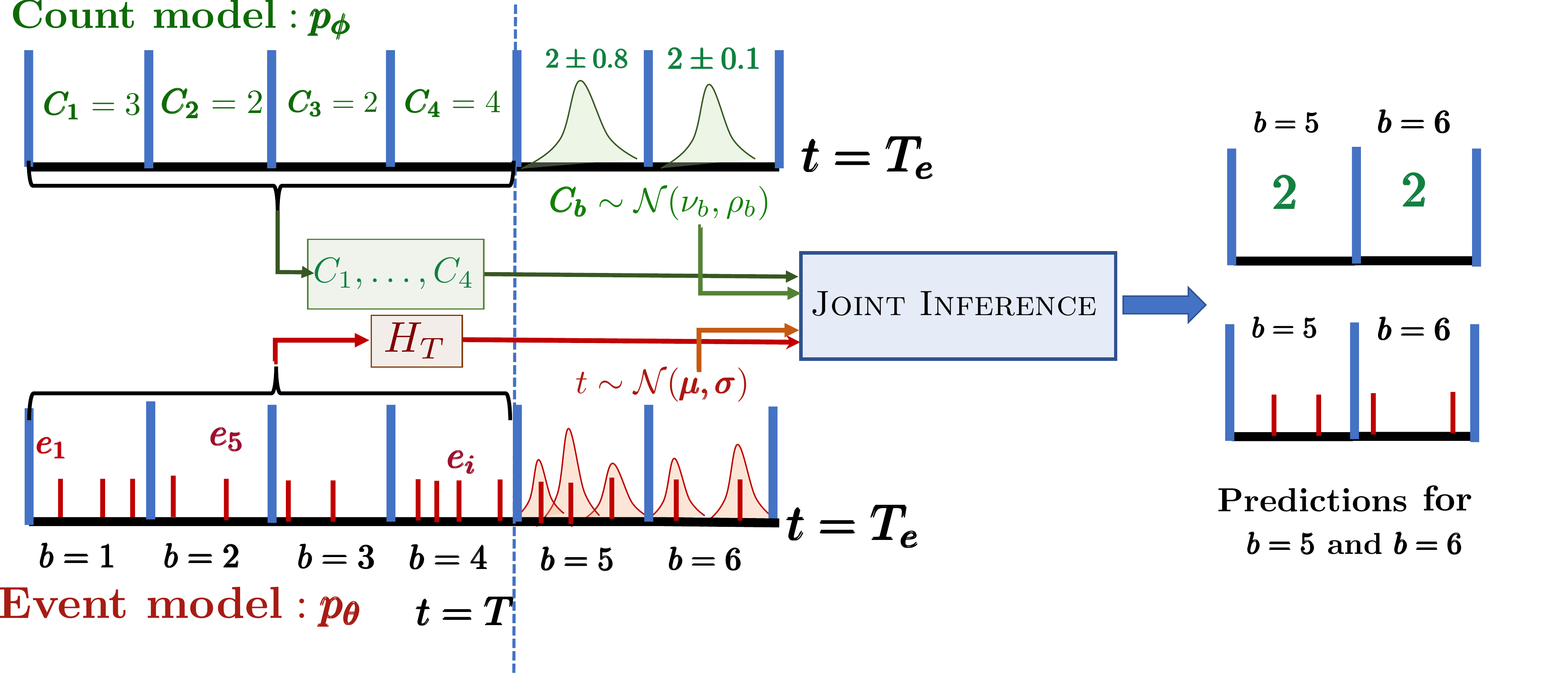}
\caption{Overview of the inference task in \our. In this example, the time horizon is split into six bins. The eleven events in the first four bins comprise the known history $\Hcal_T$.  The event-model $p_\theta$ conditioned on $\Hcal_T$ predicts five events until $\Tend$ spanning two bins. The count model at the top predicts two Gaussians of mean 2 each.  These are combined by \our's joint inference algorithm (Algorithm~\ref{alg:basic-inference}) to get the revised event predictions shown on the right. Marks are omitted for clarity.}
\label{fig:example}
\end{figure*}

\xhdr{Event model}
Our event model is a generative process which draws the next event $(m,t)$, given the history of events $\Hcal_t$. In several applications
~\cite{jing2017neural,xiao2017modeling} 
the arrival times as well as the marks of the subsequent events
depend on the history of previous events.
Therefore, we capture such inter-event dependencies 
by realizing our event model using a conditional density function $p_{\theta}(\bullet|\Hcal_t)$.
Following several existing MTPP models~\cite{du2016recurrent,mei2017neural,jing2017neural,xiao2017modeling} we model $p_{\theta}(\bullet | \Hcal_t)$
by means of a recurrent neural network with parameter $\theta$, which embeds the history of events
in compact vectors $\hb_{\bullet}$. It has three layers: (i) input layer, (ii) hidden layer and (iii) output layer.
In the following, we illustrate them in detail.

\noindent\emph{--- Input layer.} Upon arrival of the $i$-th event $e=(m_i,t_i)$, the input layer 
 transforms $m_i$ into an embedding vector $\overline{\mb}_i$ and computes the inter-arrival gap $\delta_i$, 
 which are used by next layers afterwards. More specifically, it computes
\begin{align}
    &\overline{\mb}_i = m_i \wb_m + \bb_m \label{eq:mm},\\
    & \delta_i = t_i - t_{i-1},
\end{align}
where $\wb_m$ is embedding matrix and $\bb_m$ is bias. 

\noindent\emph{--- Hidden layer.}
This layer embeds the history of events in 
the sequence of hidden state vectors $(\hb_\bullet)$ using a recurrent neural network.
More specifically, it takes three signals as input: (i) the embeddings $\overline{\mb}_i$ and
(ii) the inter-arrival time duration $\delta_i$, which is computed in the previous layer, as well
as (iii) the hour of the event as an additional feature $f_i$; and then updates the hidden state $\hb_\bullet$ using a gated recurrent unit 
as follows:
\begin{align}
    \hb_i = \text{GRU}_{\wb_h}(\hb_{i-1}; \overline{\mb}_i, \delta_i, f_i ).
\end{align}
Note that $\hb_i$ summarizes the history of first $i$ events.

\noindent\emph{--- Output layer.} Finally, the output layer computes the distribution of the  mark $m_{i+1}$ and timing $t_{i+1}$  of the next event as follows: We parameterize the distribution over marks as a softmax over the hidden states:
\begin{align}
& \PP(m_{i+1}=c)  =  \dfrac{\exp(\wb_{y,c} ^\top \hb_i + b_{y,c} )}{ \sum_{j=1} ^K \exp(\wb_{y,j} ^\top \hb_i + b_{y,j} ) } \label{eq:mark}
\end{align}
Similar to~\cite{shchur2019intensity}, we use a Gaussian distribution to model the gap $\delta_{i}$ to the next event:
\begin{align}
   & \delta_{i}  \sim \Ncal(\mu(\hb_i),\sigma(\hb_i)), \ ~~t_{i+1} = t_i +\delta_{i}\label{eq:delta-distr}. 
\end{align}
Here $\mu(\hb_i)$ the mean gap and its standard deviation $\sigma(\hb_i)$ are computed from the hidden state as:
\begin{align}
    \mu(\hb_i) & = \text{softplus}(\wb^\top _{\mu} \hb_i + b_{\mu}) \\
    \sigma(\hb_i) & = \text{softplus}(\wb^\top _{\sigma} \hb_i + b_{\sigma}).
\end{align}
Here, $\theta=\{\wb_{\bullet},b_{\bullet}\}$ are the set of trainable parameters. The Gaussian density provided more accurate long-term forecasts than existing intensity-based approaches such as RMTPP \cite{du2016recurrent}. Our ablation study in Table \ref{tab:ablation} shows this with a version of \our. 
%

The above event model via its auto-regressive structure is effective in capturing the arrival process of events at a microscopic level.
Indeed it is sufficiently capable of accurately predicting
events in the short-term future. However, its long-term predictions suffers due to cascading errors when auto-regressing on predicted events.
The count model is designed to contain this drift. 


\xhdr{Count model}
Here we aim to capture the number of events arriving in a sequence of time intervals. 
We partition time into equal time-intervals
--- called as bins---
of size $\Delta$ which is a hyper-parameter of our model (Figure~\ref{fig:example} shows an example).  Given a history of events $\Hcal_T$, we
develop a simple distribution $p_{\phi}$ which generates 
the total count of events for subsequent $n$ bins.
Let $\Ie_s$ be the time-interval $[T+(s-1)\Delta,T+s\Delta)$, and  $C_{s}$ denote the number of events occurring within it.
%
We factorize the distribution  $p_{\phi}$ over the $n$ future bins independently over each of the bins while conditioning on the known history $\Hcal_{T}$, 
and properties of the predicted bin.   
\begin{align}
 &p_{\phi}(C_{{n}}, C_{{n-1}},\cdots, C_{{1}} \,|\,\Hcal_T, I_{n},\ldots,I_{1})
 = \prod_{j=1} ^n  p_{\phi} (C_{{j} }\,|\, \Hcal_T, I_{j}) 
\end{align}
%
%

This conditionally independent model provided better accuracy than an auto-regressive model that would require conditioning on future unknown counts.  Similar observations have been made for time-series models in \cite{Deshpande2019, wen2017multi}.

Each  $p_{\phi}(C_{{j} }\,|\, \Hcal_T, I_{j})$ is modeled as a Gaussian distribution with mean $\nu_{j,\phi}$ and variance $\rho_{j,\phi}$.  
Gaussian distribution provides explicit control on variance. This is necessary for efficient inference described in Section \ref{sec:infer}. Although the domain of Gaussian distribution is $-\infty$ to $+\infty$, it is convenient for training and does not lead to any issues during inference. 
A feed-forward network with parameters $\phi$, learns these parameters as a function of features extracted from the history $\Hcal_T$ and current interval $I_{j}$ as follows:  From  a time interval  we extract time-features such as the hour-of-the-day in the mid-point of the bin. Then from $\Hcal_T$ we extract the counts of events in the most recent $n^-$ bins before $T$
and time features from their corresponding bins. 


%

\xhdr{Learning the parameters $\theta$ and $\phi$}
Given a stream of observed events $\set{e^{-} _i}$ during the time window $(0,T]$, 
we learn the event model $\theta$ by maximizing the following likelihood function:
\begin{align}
 \underset{\theta}{\text{maximize}}\quad \sum_{e_i\in\Hcal _T} \log p_{\theta}(e^- _i\,|\, \Hcal_{t_i}).\label{eq:learn1}
\end{align}
In order to train the count model, we first group the events into different bins 
of same width $\Delta$. Next, we sample them in different batches of $n^-+n$ bins and then learn $\phi$ in the following manner:
\begin{align}
\hspace{-2mm} \underset{\phi}{\text{maximize}}  \ \underset{H_\binid \sim p_{\text{Data}} }{\EE}
\sum_{j=1}^n \log p_\phi(C^- _{{\binid+j}}|H_{s}, I^- _{{\binid+j}})
\label{eq:learn2}
\end{align}
Here, $H_s$ denotes a history of events between time $(s-n^-)\Delta$ and $s$, and $C^- _{\binid+j}$ denotes the observed counts of events in bin $I_{s+j}$.

\section{Inference}
\label{sec:infer}
In this section, we formulate our inference procedure over the trained models ($p_{\etheta},p_{\ephi}$) for forecasting all events (marks and time) within a user-provided future time $\Tend$ given the history $\Hcal_T$ of events before $T < \Tend$.

\xhdr{Inference with Event-only Model} First, we review an existing method of solving this inference task using the event-only method $p_\etheta$.  Note  $p_\etheta$ is an auto-regressive model that provides a distribution on the next event $e_{i+1}$ given known historical events $\Hcal_T$ before $T$ and predicted prior events $\ee_1, \ldots \ee_{i}$, that is $p_\etheta(e_{i+1}\,|\,\Hcal_T,\ee_1, \ldots \ee_{i})$.  Let $\hb_i$ denote the RNN state  after input of events in the history $\Hcal_T$ and predicted events $\ee_1, \ldots \ee_{i}$. Based on the state, we predict a distribution of the next gap via Eq.~\ref{eq:delta-distr} and next mark using Eq.~\ref{eq:mark}.  The predicted times and marks of the next event are just the modes of the respective distribution as:  $\ee_{i+1}=(\emk_{i+1} = \argmax_m P(m_{i+1}=m), \et_{i+1} = \et_i + \mu(\hb_i))$.  The predicted event is input to the RNN to get a new state and we repeat the process until we predict an event with time $>\Tend$. 


As mentioned earlier, the events predicted by such forward-sampling method on $p_\etheta$ alone is subject to drift particularly when $\Tend$ is far from $T$. We next go over how \our\ captures the drift by generating an event sequence that jointly maximizes the probability of the event and count model.  

\xhdr{Joint Inference Objective of \our} The event model gives a distribution of the next event given all previous events whereas the count model $p_\ephi$ imposes a distribution over the number of events that fall within the regular $\Delta$-sized bins between time $T$ and $\Tend$.  For simplicity of exposition we assume that $\Tend$ aligns with a bin boundary, i.e., $\Tend = T + \nend \Delta$ for a positive $\nend$. 
During inference we wish to determine the sequence of events  that maximizes the product of their joint probability as follows:
\begin{align}
& {\underset{\substack{r, e_1,\ldots, e_r,\\ C_1,\ldots,C_{\nend}}}{\max}}     \bigg[\sum_{i=1}^{r }\log p_{\etheta}  ( e_{i} \,|\, \Hcal_{T}, e_1,\ldots,e_{i-1} ) + \sum_{b=1}^\nend \log p_{\ephi}(C_b \,|\, \Hcal_{T}, I_b) \bigg] \label{eq:inf-def}\\
 &   \text{such that, \ }  t_r < \Tend,~~~~~\big|\{e_i \,|\, t_i \in I_b\}\big| =  C_b ~~~\forall  b \in [\nend] \label{eq:con-def}
\end{align}

Unlike the number of bins, the number of events $r$ is unknown and part of the optimization process. The constraints ensure that the last event ends before $\Tend$ and there is consensus between the count and event model. Solving the above optimization problem exactly over all possible event sequences completing before $T_e$ is intractable for several confounding reasons --- the event model expresses the dependence of an event over {\em all} previous events, and that too via arbitrary non-linear functions.  Also, it is not obvious how to enforce the integral constraint on the number of events in a bin as expressed in Eq.~\ref{eq:con-def}.  

\xhdr{Tractable Decomposition of the Joint Objective} We propose two simplifications that allow us to decompose the above intractable objective into a sequence of optimization problems that are optimally solvable.
First, we decompose the objective into $\nend$ stages.  In the $b$-th stage we infer the set of events whose times fall within the $b$-th bin $I_b$ assuming we already predicted the set of all events before that bin. Call these $\Eb_b = \ee_1,\ldots, \ee_{r_b}$ where $r_b=|\Eb_b|$ denotes the number of predicted events before start of $b$-th bin, i.e, left of $I_b$. 
Second, we fix the RNN state $\hb_i$ for all potential events in $I_b$ to their unconstrained values as follows: Starting with the RNN state $\hb_{r_b}$, we perform forward sampling as in the event-only baseline until we sample an upper limit  $\Cmax$ of events likely to be in $I_b$.  
We will discuss how to choose $\Cmax$ later.  Once the RNN state $\hb_i$ is fixed, the distribution of the gap between the $i$-th and $(i+1)$th event is modeled as a Gaussian $\Ncal(\mu(\hb_i), \sigma(\hb_i))$ and the predicted mark $\emk_{i+1}$ is also fixed.
%
%
We can then rewrite the above inference problem for the events on $b$-th bin as a double optimization problem as follows:

\begin{align}
& {\underset{c \in [\Cmax]}{\max}} \bigg[ {\underset{g_1,\ldots g_c}{\max}}    \sum_{i=1}^{\Cmax} \log \Ncal(g_i;\mu(\hb_{r_b+i}), \sigma(\hb_{r_b+i})) + \log \Ncal(c; \nu_b, \rho_b)\bigg] \nonumber \\
 &   \text{such that, \ } g_i \ge 0,~~  
 \sum_{i=1}^c g_i \le \Delta, ~\sum_{i=1}^{c+1} g_i > \Delta,~~~~~\et_{r_b}+g_1 \in I_b  \label{eq:inf-tractable}
\end{align}
In the above equation, the constraints in the inner optimization just ensure that exactly $c$ events are inside bin $I_b$. All constraints are linear in $g_i$ unlike in Eq.~\ref{eq:con-def}.
The optimization problem in Eq.~\ref{eq:inf-tractable} is amenable to efficient inference:  For a fixed $c$, the inner maximization is over real-valued gap variables $g_i$ with a concave quadratic objective and linear constraints. Thus, for a given $c$, the optimal gap values can be efficiently solved using any off-the-shelf QP solver.  
The outer maximization is over integral values of $c$ but we use a simple binary search between the range 0 and $\Cmax$ to solve the above in $\log(\Cmax)$ time.

Let $c^*,g^*_1,\ldots,g^*_{c^*}$ denote the optimal solution.  Using these we expand the predicted event sequence from $r_b$ by $c^*$ more events as $(\emk_{r_b+1}, \et_{r_b}+g^*_1),\ldots (\emk_{r_b+c^*}, \et_{r_b}+\sum_{i=1}^{c^*} g^*_i)$.  We append these to $\Eb_b$ to get the new history of predicted events $\Eb_{b+1}$ conditioned on which we predict events for the $(b+1)$-th bin.  The final set of predicted events are obtained after $n_e$ stages in $E_{n_e+1}$

\xhdr{Choosing $\Cmax$ }
Let $C_E$ denote the count of events in bin $I_b$ when each gap $g_i$ is set to its unconstrained optimum value of $\mu(.)$. We obtain this value as we perform forward sampling from RNN state $\hb_{r_b}$.  The optimum value of $c$ from the count-only model is $\nu_b$.  Due to the unimodal nature of the count model $p_{\ephi}$,  one can show that the optimal $c^*$ lies between $\nu_b$ and $C_E$. Thus, we set the value $\Cmax = \max(\nu_b+1, C_E)$.  Also, to protect against degenerate event-models that do not advance time of events, we upper bound $\Cmax$ to be $\nu_b+\rho_b$ since the count model is significantly more accurate, and the optimum $c^*$ is close to its mode $\nu_b$.

%

%

\setlength{\textfloatsep}{0pt}
\begin{algorithm}[t]                    
 	\small
	\caption{Inference of events in the $[T,\Tend)$}
	\label{alg:basic-inference}
	\begin{algorithmic}[1]
	\STATE \textbf{Input: } Trained event model and trained count model $p_{\etheta},\ p_{\ephi}$, event history $\Hcal_T$, end time $\Tend= T+n_e\Delta$.
	\STATE \textbf{Output: }  Forecast events $\{\ee\given \et\in [T,\Tend) \}$
 
    \STATE $\Eb \leftarrow \emptyset$  \texttt{/* Predicted events so far */}
	\FOR{$b$ in $[n_e]$}
		\vspace{1.3mm}
	    \STATE $\nu_b,\rho_b \leftarrow$ Count distribution from $p_\phi(.|\Hcal_T, I_b)$ 
	    \STATE $\hb,\Cmax \leftarrow \textsc{RNNStates}(p_\theta, \Hcal_T, \Eb, \nu_b, b)$ 
	     \texttt{/*set $\hb_{\bullet}$ */}
	    \vspace{1.3mm}
		\STATE \texttt{\small /* Solve the optimization problem in Eq.~\ref{eq:inf-tractable}*/}
	    \STATE $\Eb \leftarrow  \Eb + \textsc{OptimizeInBin}(\hb, \nu_b,\rho_b,\Cmax, I_b) $
	\ENDFOR
	\STATE \text{Return} $\Eb$
	\end{algorithmic}
\end{algorithm}

\xhdr{Overall Algorithm}
Algorithm~\ref{alg:basic-inference} summarizes \our's inference method. An example is shown in Figure~\ref{fig:example}. To predict the events in the $b$-th bin, we first invoke the count model $p_\phi$ and get mean count $\nu_b$, variance $\rho_b$.
We then forward step through the event RNN  $p_\theta$ after conditioning on previous events $\Hcal_T, \Eb$.  We then continue forward sampling until bin end or $\mu_b+1$, and return the visited RNN states, and number of steps $\Cmax$. Now, we invoke the optimization problem in Eq.~\ref{eq:inf-tractable} to get 
the predicted events in the 
$b$th bin which we then append to $\Eb$.  

\section{Experiments}
\label{sec:expts}
In this section, we evaluate our method against five state-of-the-art existing methods, on four real datasets.

\subsection{Datasets}
We use four real world datasets that contain diverse characteristics in terms of 
their application domains and temporal statistics. We also summarize the details of these datasets in Table~\ref{tab:dataset}.

\xhdr{Election}~\cite{de2018demarcating} This dataset contains  tweets related to presidential election results in the United-States, collected from 7th April to 13th April, 2016. Here, given a tweet $e$, the mark $m$ indicates the user who posted it and the time $t$ indicates the time of the post. 

\xhdr{Taxi}~\cite{taxi} This contains the pickup, drop-off timestamps and pickup, drop-off locations of taxis in New York city from 1st Jan 2019 to 28th Feb 2019. The dataset is categorized by zones. 
In our experiments we only consider pick up zone with zone id $237$. We consider each travel as an event $e=(m,t)$,  
with pick up time denoted by $t$ and drop-off zone as the marker $m$.


\xhdr{Traffic-911}~\cite{emergency-911}
This dataset consists of emergency calls related to road traffic in the US, in which each event  contains timestamp of the call and location of the caller, which we treat as a marker. 

\xhdr{EMS-911}~\cite{emergency-911}
This dataset consists of emergency calls related to medical services in the US, in which each event  contains timestamps of the call, and location of the caller which we treat as the marker. 

For all datasets, we rank markers based on their occurrence frequency and keep the top $10$ markers. Rest of the markers are merged into a single mark. Hence, we have $11$ markers in each dataset.

\begin{table}[t]
\resizebox{0.47\textwidth}{!}{
\begin{tabular}{|l|r|r|r|r|l|}
\hline
Dataset     & Train  & $\EE[t]$ & $\sigma[t]$  & Avg. \#Events & Bin Size     \\
     &  Size  &     &   & in $[T, T_e)$ & ($\Delta$)      \\
\hline
Elections       & 51859         & 7.0       & 5.8   & 203       & 7 mins.\\
\hline
Taxi       &   399433         & 8.0       & 25.8    &   1254   & 1 hour      \\
\hline
Traffic-911 &   115463         & 778       & 1517   &  281    & 1 day     \\
\hline
EMS-911   &   182845         & 492       & 601  &   275    & 12 hours     \\
\hline
\end{tabular}}
\caption{Statistics of the datasets used in our experiments. Train Size denotes the number of events in the training set. $\EE[t]$ and $\sigma[t]$ denote the mean and variance of the inter-event arrival time.}
\label{tab:dataset}
\end{table}

\subsection{Methods Compared}
\label{sec:expt:baselines}
We compare \our\ against five other methods spanning a varied set of loss functions and  architectures: The first two (RMTPP, THP) are trained to predict the next event via intensity functions using maximum likelihood (Sec~\ref{sec:overviewMTPP}).  The next two (WGAN and Seq2Seq) are trained to predict a number of future events using a sequence-level Wasserstein loss and are better suited for long-term forecasting.  The last uses a two-level hierarchy to capture long-term dynamics.   We present more details below:

\xhdr{RMTPP}
RMTPP \cite{du2016recurrent} is one of the earliest neural point process model that uses a three layer recurrent neural network
to model the intensity function and mark distribution of an MTPP.  

\xhdr{Transformer Hawkes Process (THP)}
THP \cite{zuo2020transformer} is more recent and uses Transformers~\cite{Vaswani2017} instead of RNNs to model the intensity function of  the next event. 
The THP leverages the positional encoding in the transformer model to encode the timestamp. 
%
%

\xhdr{WGAN}
Wasserstein TPPs  \cite{xiao2017wasserstein} train a generative adversarial network to generate an event sequence. 
A Homogeneous Poisson process provides the input noise to the generator of future events, which by a Wasserstein discriminator loss is trained to resemble real events.
%
%
Since our predicted events are conditioned on the input history, we initialize the generator by encoding known history of events using an RNN. 


\xhdr{Seq2Seq}
is a conditional generative model~\cite{xiao2018learning}, in which, an encoder-decoder model for sequence-to-sequence learning is trained by maximizing the likelihood of the output sequence. Also added is a Wasserstein loss computed via a CNN-based discriminator. 

\xhdr{Hierarchical Generation} We designed this method to explore if hierarchical models~\cite{wavenets,vassoy2019time,borovykh2017conditional}, could be just as effective as our count-model to capture macroscopic dynamics.  
We create a two-level hierarchy where the top-level events are 
compound events of $\tau$ consecutive events.  We train a second event-only model $p_{\psi}(\bullet|H_t)$ over the compound events to replace the count-model.
%
%
Using trained models ($p_{\etheta}, p_{\psi}$) we perform inference similar to Eq.~\ref{eq:inf-tractable}. However, since compound model $p_{\psi}$ imposes a distribution over every $\tau$-th event, we solve the following optimization problem for every $j$-th compound event:
\vspace{-3mm}
\begin{align}
	\nonumber {\underset{g_1\ldots g_{\tau}, g_i \in \RR^+}\max} \bigg[ & \sum_{i=1}^{\tau} \log \Ncal(g_i;\mu(\hb_{j\tau+i}), \sigma(\hb_{j\tau+i})) ~~+ ~~~~ \\[-1em]
	& \log \Ncal(\sum_{i=1}^{\tau} g_i; \mu(\hb_{j}^c), \sigma(\hb_{j}^c)) \bigg] \label{eq:hir-inf-tractable}
\end{align}
Similar to Eq.~\ref{eq:inf-tractable}, the maximization is over positive real-valued gap variables $g_i$ and with a concave quadratic objective. Here, the number of stages is not fixed to $\nend$, but we stop when the last predicted time-stamp is greater than $\Tend$.

\subsection{Evaluation protocol}
We create train-validation-test splits for each dataset by selecting the first 60\% time-ordered events as training set, next 20\% as validation and rest 20\% as test set. We chose the value of the bin-size $\Delta$ so that each bin has at least five events on average while aligning with standard time periodicity as shown in Table~\ref{tab:dataset}.  
A test `instance' starts at a random time $T_s$ within the test time, includes all events up to $T=T_s+20\Delta$ as the known history $\Hcal_T$, and treat the interval between $T$ and $\Tend=T+3\Delta$ as the forecast horizon.  The average number of events in the forecast horizon ranges between 200 and 1250 across the four datasets (shown in Table~\ref{tab:dataset}).
For training the count model $p_\phi$ we created instances using the same scheme.  The event model $p_\theta$ just trains for the next event using random event sub sequences of length 80.

\xhdr{Architectural Details}
For event model, we use a single layer recurrent network with GRU cell of $32$ units. We fixed the batch size to $32$ and used Adam optimizer with learning rate $1\mathrm{e}{-3}$. The size of the embedding vector of a mark is set to $8$.  We train the event model for $10$ epochs. We checkpoint the model at the end of each epoch and select the model that gives least validation error.
The Count model is a feed-forward network with three hidden layers of $32$ units, all with ReLU activation. The input layer of count model has $40$ units, corresponding to counts of $20$ input bins and hour-of-day at the mid-point of each bin. The output layer predicts the Gaussian parameters $\nu_j, \rho_j$ for each future bin $j$.

\begin{figure}
    \centering
    \includegraphics[width=0.235\textwidth]{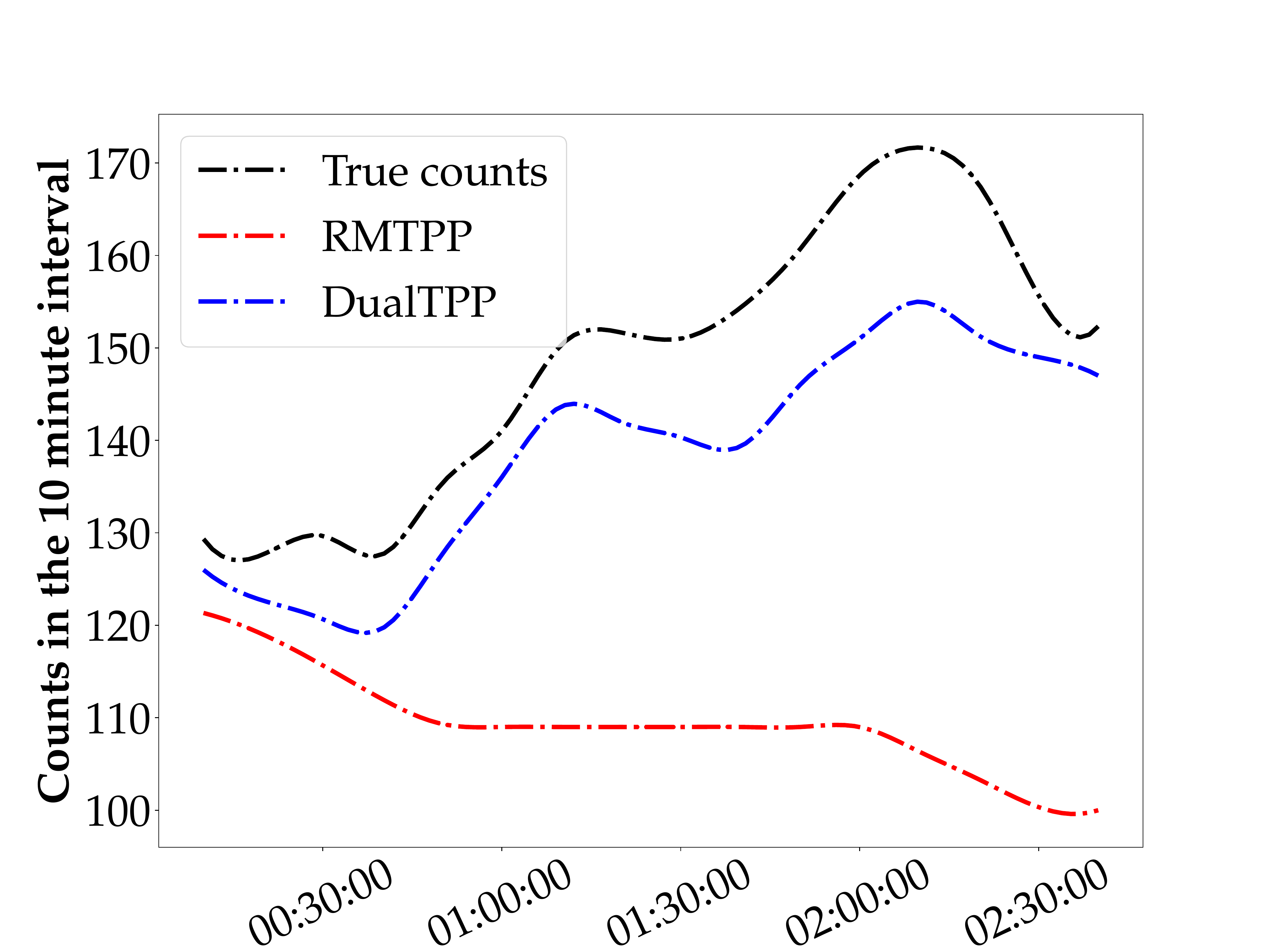}
    \includegraphics[width=0.235\textwidth]{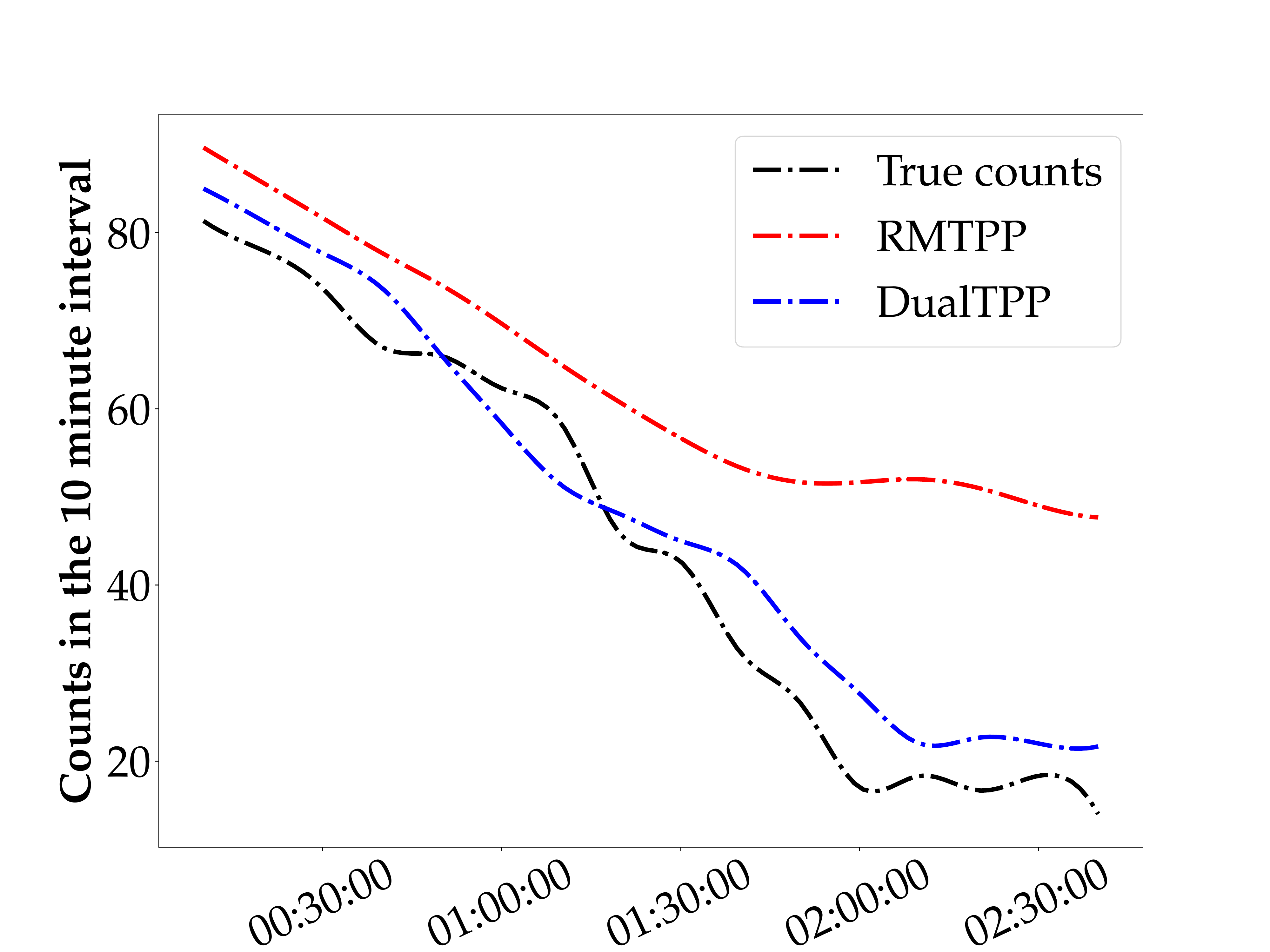}
    \includegraphics[width=0.235\textwidth]{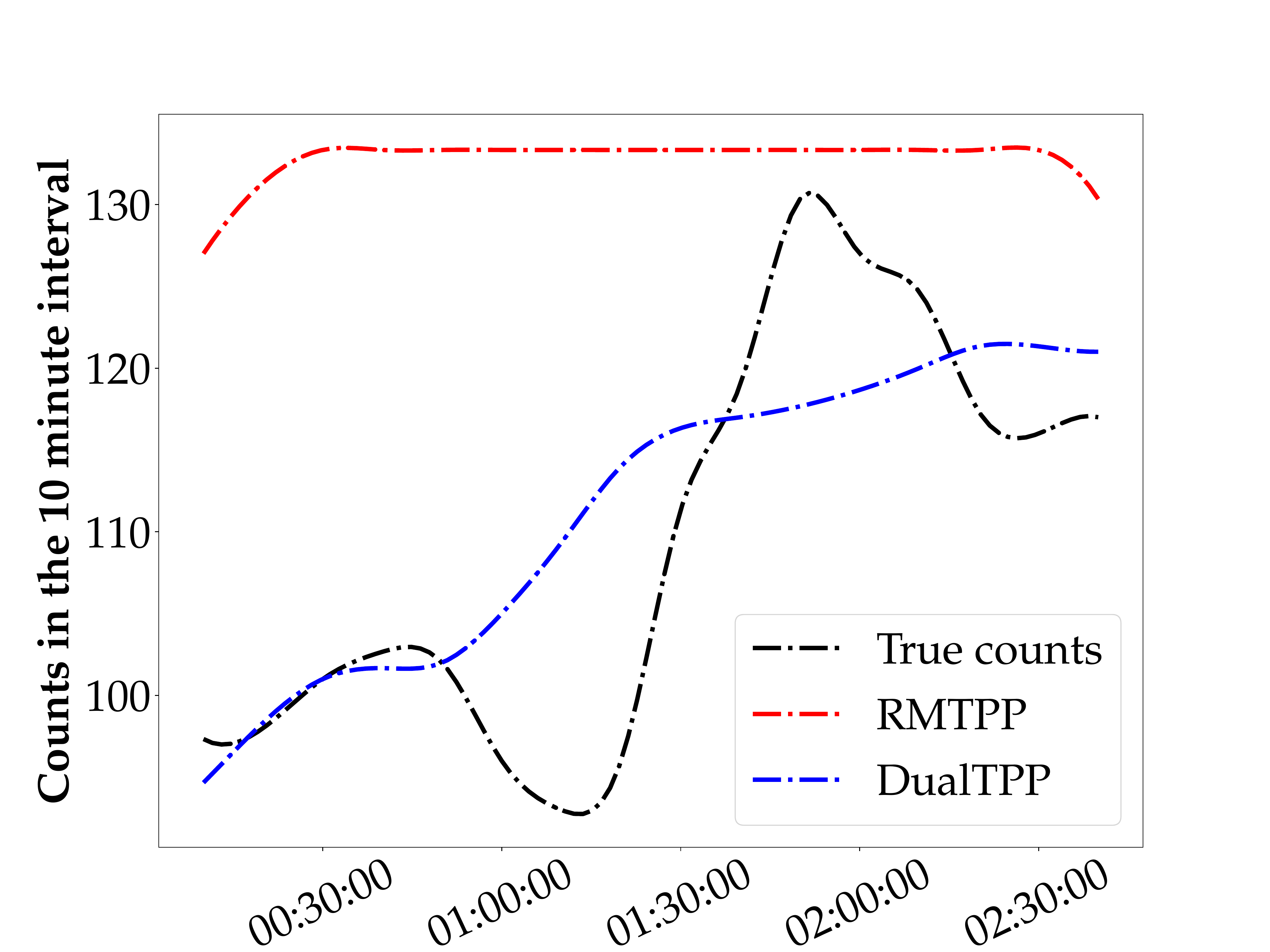}
    
    
    \caption{Anecdotal examples of variation of counts against time, collected from Taxi datasets. 
    They show that \our can mimic the high level trajectory more accurately than RMTPP. In the second example, we observe that RMTPP and \our show similar nowcasting performance, whereas \our shows more accurate forecasting performance than RMTPP.}
    \label{fig:Anecdotes}
\end{figure}

\xhdr{Evaluation Metrics}
We use three metrics to measure performance.
%
First, we measure the Wasserstein distance between predicted and actual event sequences to assess the microscopic dynamics between events. 
Given true event times $\Hcal :=\{t_1,\ldots,t_{|\Hcal|}\}$ 
in an interval $[\Tst,\Tend)$
and the corresponding predicted events $\widehat{\Hcal} :=\{\et_1,\ldots,\et_{|\widehat{\Hcal}|} \}$, assuming without loss of generality, $|\Hcal|<|\widehat{\Hcal}|$, we compute the Wasserstein distance\footnote{Here, the term `Wasserstein distance' is overloaded. However, as shown in \cite{xiao2017wasserstein}, for distributions with point masses, Wasserstein distance simplifies to Eq.~\ref{Eqn:wassdist}.}~\cite{xiao2017wasserstein} between the two sequence of events as
\begin{align}
 \label{Eqn:wassdist}
 \text{WassDist}(\Hcal, \widehat{\Hcal}) = \sum_{i=1}^{|\Hcal|} |t_i - \et_i| + 
 \sum_{i=|\Hcal|+1}^{\widehat{\Hcal}} (\Tend  -  \et_{i})
\end{align}
We  randomly sample several such intervals $[\Tst,\Tend)$ and report the 
average of  \text{WassDist} of all intervals.
Second, to assess the macroscopic modeling component of each method we define a CountMAE that aims to measure the relative error in predicted count in randomly sampled time interval:  
\begin{align}
    \text{CountMAE} =  \frac{1}{M}\sum_{i\in M}\dfrac{\big|\set{e\,|\,t \in \Ical^{(i)}}\big| - \big|\set{\ee\,|\,\et \in \Ical^{(i)}}\big|}{\big| \set{e\,|\,t \in \Ical^{(i)}} \big|},
\end{align}
where $\Ical^{(i)}$ is randomly sampled in test-horizon and we sample $M$ such intervals.
Finally, for evaluating accuracy of the predicted discrete mark sequence, we compare our generated mark sequence with the true mark sequence (which could be of a different length) using the BLEU score popular in the  NLP community~\cite{bleu}.
\begin{table}
\small
\begin{tabular}{|l|l|r|r|r|}
 \hline
Dataset     & Model               & Wass. & BLEU  & Count \\
            &                     & dist  & Score & MAE   \\ 
\hline
Elections       & \RmtppNllSimu~\cite{du2016recurrent}       & 1231      & 0.684      & 26.7      \\
            & \Transformer~\cite{zuo2020transformer}        & 1458      & 0.579      & 31.8      \\
            & \WGAN~\cite{xiao2017wasserstein}               & 442       &   -         & 10.0        \\
            & \SeqSeq~\cite{xiao2018learning}             & 739       &     -       & 15.9      \\
            & \RmtppMseOptComp    & 415       & 0.880       & 8.5       \\
            & \RmtppMsevarOpt     & \textbf{267}       & \textbf{0.882}      & \textbf{5.0}     \\ \hline
Taxi        & \RmtppNllSimu~\cite{du2016recurrent}       & 9826      & 0.089      & 288     \\
            & \WGAN~\cite{xiao2017wasserstein}               & 4060      &   -         & 128     \\
            & \SeqSeq~\cite{xiao2018learning}             & 5105      &       -     & 161       \\
            & \RmtppMsevarOptComp & 8838      & 0.088      & 206     \\
            & \RmtppMsevarOpt     & \textbf{1923}      & \textbf{0.090}       & \textbf{39}      \\ \hline
Traffic-911 & \RmtppNllSimu~\cite{du2016recurrent}       & 2406      & \textbf{0.248}      & 41.7      \\
            & \Transformer~\cite{zuo2020transformer}        & 6096      &    0.081       & 110.0      \\
            & \WGAN~\cite{xiao2018learning}               & 3892      &         -   & 69.0        \\
            & \SeqSeq~\cite{xiao2018learning}             & 4520      &          -  & 83.0        \\
            & \RmtppMseOptComp    & 1853      & 0.211      & 33.1      \\
            & \RmtppMseOpt        & \textbf{1700}      & 0.221      & \textbf{29.1}      \\
            \hline
EMS-911     & \RmtppNllSimu~\cite{du2016recurrent}       & 2674      & 0.162      & 20.9      \\
            & \Transformer~\cite{zuo2020transformer}        & 5792      & 0.070       & 50.0        \\
            & \WGAN~\cite{xiao2018learning}               & 2432      &      -      & 19.3      \\
            & \SeqSeq~\cite{xiao2018learning}             & 9856      &      -      & 90.3      \\
            & \RmtppMseOptComp    & 1639      & \textbf{0.163}      & 11.8      \\
            & \RmtppMseOpt        & \textbf{1419}      & \textbf{0.163}      & \textbf{10.1}     \\ \hline
\end{tabular}
\caption{Comparative analysis of our method against all baselines across all datasets in terms of WassDist, BLEUScore, and CountMAE.  It shows the \our consistently outperforms all the baselines.}
\label{tab:main}
\end{table}
%
%
\subsection{Results}
In this section, we first compare \our against the five methods of Sec~\ref{sec:expt:baselines}, and then analyze how accurately it can forecast events in a distant time. Next, we provide a thorough ablation study on \our. 

\xhdr{Comparative analysis}
Here we compare \our against five state-of-the-art methods. 
\WGAN and \SeqSeq papers do not model marks, hence their BLEU scores are omitted.
Table~\ref{tab:main} summarizes the results, which reveals the following observations.  

\begin{itemize}[leftmargin=*]
\item[(1)] \our achieves significant accuracy gains beyond all five methods, in terms of all three metrics \ie, CountMAE, WassDist and BLEUScore. For some datasets, e.g. Taxi the gains by our method are particularly striking --- our error in counts is 
39, 
and the closest alternative has almost 
three 
times higher error! Even for microscopic inter-event dynamics as measured by the Wasserstein distance we achieved a factor of two reduction. Figure~\ref{fig:Anecdotes} shows three anecdotal sequences comparing counts of events in different time-intervals of \our (Blue) against actual (Black) and the RMTPP baseline (red).  Notice how RMTPP drifts away whereas \our tracks the actual.
%
\item[(2)] The \RmtppMsevarOptComp\ variant of our method is the second best performer, but its performance is substantially poor compared to \our, establishing that the alternative count-based perspective is as important as viewing events at different scales for accurate long-term perspective. More specifically,  the \RmtppMseOptComp\ variant considers aggregating a fixed number of events, which makes it oblivious to the prediction for heterogeneous counts in an arbitrary time interval. \our aims to overcome these limitations by means of both the event and the count model, which characterize both short term and long term characteristics of an event sequence.
\item[(3)] Both RMTPP and \Transformer\ are much worse than \our. \WGAN 
 and \SeqSeq provide unreliable performance and show large variance across datasets.
\end{itemize}
\input{per_bin_results}
\input{sensitivity}

\xhdr{Performance on long term forecasting} 
Next we analyze the performance difference further by looking at errors in different forecast time-intervals in the future in
Figure~\ref{fig:forecasting}.  Here, on the X-axis each tick gives the average number of events since the known history $T$ in gold and on the Y-axis we show the Wasserstein distance for events predicted between time of two consecutive ticks. We observe the expected pattern that events further into the future have larger error than closer events for all methods.  
However, \our shows a modest deterioration, whereas,
both RMTPP and \RmtppMsevarOptComp show a significant deterioration.
For example in the leftmost plots on the Election dataset, the Wasserstein distance for the first 68 events increases from 500 to almost 800 for RMTPP, but only from 200 to 270 for \our.
%

We measure sensitivity of our results to bin-sizes by varying the bin-size $\Delta$, and correspondingly the 
forecast horizon 
$[T, \Tend=3\Delta]$. %
Figure~\ref{fig:sensitivity} shows the Wasserstein distance between true and predicted events across different bin sizes  on the Taxi dataset. We find that \our continues to perform better than competing methods across all bin sizes.  


    





\xhdr{Ablation Study}
We perform ablation study using variants of \our to analyze which elements of our design contributed most to our observed gains.  We evaluate these variants using the Wasserstein distance metric and summarize in Table~\ref{tab:ablation}.

First we see the performance obtained by our Event-only model.   We observe that the event-only model performs much worse than \our, establishing the importance of the count model to capture its drift.
We next compare with the \CountOnly model, where we first predict counts of event for the $b$-th bin ($\nu_b$), and then randomly generate $\nu_b$ events in the $b$-th bin.  In this case, marks are ignored.  We observe that the count-only model is worse than \our\ but it performs much better than the Event-only method.  

Next, we analyze other finer characteristics of our model. In \our, the event model uses the Gaussian density whereas most existing TPP models (e.g. RMTPP and THP discussed earlier) use an intensity function.  We create a version of  \our called \RmtppNllOpt where we model the distribution $p_{\etheta}(\bullet|H_t)$ using the conditional intensity of RMTPP.  Comparing the two methods we observe that the choice of Gaussian density also contributes significantly to the gains observed in \our.



In the \MuOnlyOpt model, we predict the events in the $b$-th bin by solving the inner optimization problem in Eq.~\ref{eq:inf-tractable} only for the mean $\nu_b$, thereby treating $p_\phi$ as a point distribution. We observe a performance drop highlighting the benefit of modeling the uncertainty of the count distribution.

\begin{table}[]
\small
\begin{tabular}{|l|l|r|}
 \hline
Dataset     & Model                & Wass dist \\  \hline
Election    & \RmtppMsevarOpt      & 267       \\
            & \RmtppMseSimu        & 633       \\
            & \CountOnly           & 310       \\
            & \RmtppNllOpt         & 271       \\
            & \MuOnlyOpt           & 272       \\ \hline
Taxi        & \RmtppMsevarOpt      & 1923      \\
            & \RmtppMsevarSimu     & 5679      \\
            & \CountOnly           & 1923      \\
            & \RmtppNllOpt         & 1790      \\
            & \MuOnlyOpt           & 1916      \\ \hline 
Traffic-911 & \RmtppMseOpt         & 1700      \\
            & \RmtppMseSimu        & 1767      \\
            & \CountOnly           & 2098      \\ 
            & \RmtppNllOpt         & 2211      \\
            & \MuOnlyOpt           & 1746      \\ \hline
EMS-911     & \RmtppMseOpt         & 1419      \\
            & \RmtppMseSimu        & 1485      \\
            & \CountOnly           & 2186      \\ 
            & \RmtppNllOpt         & 2318      \\
            & \MuOnlyOpt           & 1423      \\  \hline
\end{tabular}
\caption{Ablation Study: Comparison of \our and its variants in terms of Wasserstein Distance between true and predicted events.}
\label{tab:ablation}
\end{table}

\section{Conclusions}
In this paper, we propose \our, a novel MTPP model specifically designed for long-term forecasting of events. It consists of two components--- Event-model which  captures dynamics of the underlying MTPP in a microscopic scale and Count-model which captures the macrocopic dynamics. Such a model demands a fresh approach for inferring future events. 
We design a novel inference method that solves a sequence of efficient constrained quadratic programs to achieve consensus across the two models. 
Our experiments show that \our achieves substantial accuracy gains beyond five competing methods in terms of all three metrics: Wasserstein distance that measures microscopic inter-event dynamics, CountMAE that measures macroscopic count error, and BLEU score that evaluates the sequence of generated marks. Future work in the area could include capturing other richer aggregate statistics of event sequences. Another interesting area is providing inference procedures for answering aggregate queries directly.



\bibliographystyle{abbrvnat}
\bibliography{refs,mining2,pubs,ML}

\if{0}
\newpage
\appendix

\section{Inference Ver 1}
Let $\ee_i=(\emk_i,\et_i)$ denote the prediction of the $i$-th event $e_i$ after $T$.  Thus, our goal is to predict all events $\ee_i=(\emk_i,\et_i)$ such that $T < \et_i \le \Tend$.

and  $\eC_s$ to be the prediction of $C_s$--- the count of events in the interval $((s-1)\Delta, s\Delta]$. Moreover, by slightly overloading the notation, we indicate $\eC_{[\Tst,\Tend)}$ as the predicted number of events in $[\Tst,\Tend)$.
In a sharp contrast to the prior works which deploy expensive sampling procedures using the underlying (event) model, we cast the inference task as  sequence of constrained optimization problem.  

A key dilemma with the proposed modeling framework is that the number of events generated by our trained event model $p_{\etheta}$ and the trained count model $p_{\ephi}$ need not be the same whereas, in principle, they should generate same the number of events.  
\begin{align}
\big|\bigl\{\et_i \in [\Tst,\Tend) | (\emk_i,\et_i) \sim p_{\etheta}(\bullet|\Hcal_{T})\bigr\}\big| =  \eC_{[\Tst,\Tend)} \label{eq:coherence}
\end{align}

where $ \eC_{[\Tst,\Tend)} \sim p_{\ephi}$ is the count of events in $[\Tst,\Tend)$, drawn $p_{\ephi}$. 
%
In order to establish this coherence during the forecasting task, we formulate our inference problem as follows:
\begin{align}
& {\underset{C, \set{\ee_i} }{\text{maximize}}}     \bigg[\sum_i \log p_{\etheta}  ( \set{\ee_i} \,|\, \Hcal_{T} ) + \log p_{\ephi}(C_{[\Tst,\Tend)} = C\,|\, \Hcal_{T}) \bigg] \label{eq:inf-def-ver1}\\
 &   \text{such that, \ }  \big|\{\et_i \in [\Tst,\Tend)\}\big| =  C \label{eq:con-def-ver1}
\end{align}
However, the presence of the constraint in Eq.~\ref{eq:con-def-ver1} leads the above optimization problem intractable
similarly as in training. Therefore, we resort to design heuristics in a principled way, which approximately solves the
above problem. 

\subsection{Count coherent inference procedure}
Given that the training set $[0,T)$ consists of $M$ bins $\set{\Ie_s = [(s-1)\Delta,s\Delta) | s\in [M]}$,
we first stride  the time-interval $[T-r\Delta,\infty)$ in $n$ steps and create a sequence $\set{v_1, v_2,\cdots,v_b,\cdots}$ in which, each $v_{\bullet}$ consists of $r+n$ bins, each of width $\Delta$, as given below:

\begin{align}
& v_1: \ \underbrace{\Ie_{M-r+1}, \Ie_{M-r+2},\cdots, \Ie_{M}}_{\text{Input sequence}}, 
\underbrace{\Ie_{M+1},\Ie_{M+2},\cdots, \Ie_{M+n}}_{\text{Output sequence}}\nn\\ 
&  v_2: \ \underbrace{\Ie_{M+n-r+1},\cdots, \Ie_{M+n}}_{\text{Input sequence}}, \underbrace{\Ie_{M+n+1},\cdots, \Ie_{M+2n}}_{\text{Output sequence}} \nn\\
&\qquad\qquad\qquad \qquad\qquad  \vdots\nn\\
&  v_b: \ \underbrace{\Ie_{M+(b-1)n-r+1}, \cdots, \Ie_{M+(b-1)n}}_{\text{Input sequence}}, \underbrace{\Ie_{M+(b-1)n+1}\cdots, \Ie_{M+ b\cdot n}}_{\text{Output sequence}} \nn\\
&\qquad\qquad\qquad \qquad\qquad  \vdots\label{eq:seq-intervals}
\end{align}
In order to forecast the events during time-interval $[\Tst,\Tend)$, we first identify the sequence 
$v_b$ for which the output interval  $\Ie_{M+(b-1)n+1}\cup \Ie_{M+(b-1)n+2}\cup \ldots \cup \Ie_{M+b\cdot n}$
overlaps with $[\Tst,\Tend)$.
%
Then we search for a candidate count $\eC_{M+ (b-1) n +j}$ for each such interval $\Ie_{{M+ (b-1) n +j}}$, which would maximize the underlying likelihood of events over the trained model.
%
Here, we search $\eC_{\bullet}$ within a confidence set for each bin. To this end,
we provide two types of search-based inference methods.
In the first variant we carry out a search over each possible count in the confidence set  whereas, the second variant carries out a binary search over it.

\xhdr{Grid-search based inference method}
Given the specified time interval $(\Tst,\Tend]$, let the corresponding  output interval in Eq.~\ref{eq:seq-intervals}  be $\Ie_{M+(b-1)n+1}\cup \Ie_{M+(b-1)n+2} \cup \ldots \cup \Ie_{M+b\cdot n}$ which overlaps with $(\Tst,\Tend]$. Furthermore, let us denote
$\ee_i$ to be the prediction for $i$-th event in the entire time-window $\Ie_{M+(b-1)n+1}\cup \Ie_{M+(b-1)n+2}  \cup \ldots \cup \Ie_{M+b\cdot n}$.
Next, we aim to estimate $\emk_i$ and $\et_i$ for each possible event $\ee_i$,  which maximize
their joint likelihood subject to the coherence between the
number of events generated by the event model and the count model in Eq.~\ref{eq:coherence}.

To that aim, we start with the bin $\Ie_{M+(b-1)n+1}$, predict the count
$\eC_{{M+(b-1)n+1}}$ and the events $\set{ \ee_{j} \given 1\le j \le \eC_{{M+(b-1)n+1}} }$; 
then, based on these estimates, we predict the count $\eC_{{M+(b-1)n+2} } $ and the events $\set{\ee_j \given \eC_{{M+(b-1)n+1}} +1 \le j \le \eC_{{M+(b-1)n+2}}  +  \eC_{{M+(b-1)n+1}} }$ corresponding to the second bin
$\Ie_{M+(b-1)n+2}$ and continue doing the same upto $n$ bins. In particular, given the  predictions $\set{\ee_j \given 1 \le j\le \sum_{s\in [i-1]}\eC_{{M+(b-1)n+s}}}$ upto $i-1$ bins, we first 
construct a confidence set $[C_{\min}, C_{\max}]$ and then
solve the following optimization problem for each $C\in [C_{\min}, C_{\max}]$
        \begin{align}
            \label{eq:opt-model-g}
       \hspace{-1mm}   
        \Mcal_{C}= \hspace{-4mm} \max_{\substack{\et' _1,\ldots,\et' _{C_{\max}} \\ \emk' _1,\ldots,\emk' _{C_{\max}}} }& \log p_{\etheta}  \left( \set{(\emk' _i,\et' _i)}_{i\in [C_{\max}]} \,|\, \Hcal_{ (M+(b-1)n+i-1)\Delta } \right)  \\
               \hspace{-4mm} \text{such that,} \ & \et _{C} \le (M+(b-1)n+i)\Delta  \label{eq:con1}\\ 
              \hspace{-4mm}  & \et'  _{C+1} \ge (M+(b-1)n+i)\Delta \label{eq:con2},
        \end{align}
where  $\ee' _j = (\emk'_j,\et' _j)$ predicts $j^{\text{th}}$ event in the $i$-th bin $\Be_{M+(b-1)n+i}$ for $j\le C$ \ie, 
\begin{align}
    \ee'_j = \ee_{j+ \sum_{s\in[i-1]} \eC_{M+(b-1)n+s}};
\end{align}
the constraints in Eqs.~\ref{eq:con1} and~\ref{eq:con2} make sure that
exactly $C$ events fall in the specified interval $\Be_{M+(b-1)n+i}$, which in turn satisfies the coherence between the event and count models; and,
\begin{align}
 p_{\etheta}& ( (\emk' _1,\et' _1),\ldots,(\emk' _ {C_{\max}}, \et' _{C_{\max}}) \,|\, \Hcal_{ (M+(b-1)n+i-1)\Delta} )\nn\\
 & = \sum_{j=1} ^{C_{\max}} \log p_{\etheta}( \ee' _j  \given \Hcal_{ \et_j}). \nonumber
\end{align}
Note that the optimization problem~\eqref{eq:opt-model-g}-\eqref{eq:con2} computes a \emph{restricted mode}
of the trained distribution $p_{\etheta}$ in which the first $C$ events fall in the $i$-th interval $\Be_{M+(b-1)n+i}$. 
Finally, we choose the optimal count for the $i-$th bin $\eC'_{M+(b-1)n+i}=C^{*}$ which satisfies the joint likelihood of estimated event and count models, \ie,
\begin{align}
\hspace{-2mm}
  C^{*} & = \argmax_{C}  \bigl( \Mcal_{C} + p_{\ephi}(  C_{{M+(b-1)n+i}}\hspace{-1mm}  = C| \Hcal_{(M+(b-1)n+1)\Delta} )\bigr)\label{eq:Cstar},
\end{align}
and therefore, the predicted events are given by the first $C^{*}$ events provided by the optimal solution of~\eqref{eq:opt-model-g}, which are  \todo{Do we need separate variables $(\emk_j^{*},\et_j^{*})$ to denote the optimal solution?}$\set{(\emk'_j,\et' _j)| j\in [C^{*}]}$. 
Moreover, we would like to point out that the optimization problem in Eq.~\ref{eq:opt-model-g} is a constrained quadratic program in $\set{\et_i}$ thanks to the Gaussian distribution in Eq.~\ref{eq:delta-distr}, which allows us to adopt any standard tool to efficiently solve it. On the other hand, the softmax distribution over the marks allows to predict the event by computing the maximum probability 
over all $K$ marks. Finally, among all the events predicted within all $n$ intervals, we record those which fall within the specified 
time interval $[\Tst, \Tend)$.

\setlength{\textfloatsep}{0pt}
\begin{algorithm}[t]                    
	\caption{Count coherent inference method}
	\label{alg:basic-inference}
	\begin{algorithmic}[1]
	\STATE \textbf{Input: } Trained event model and trained count model $p_{\etheta},\ p_{\ephi}$ and the specified time interval $[\Tst,\Tend)$.
	\STATE \textbf{Output: }  Forecast events $\{\ee_i\given \et_i\in [\Tst,\Tend) \}$
	%
	\STATE \texttt{/* Set search method*/}
    \STATE  \textbf{Search method:} $\texttt{search} \in\set{\texttt{Grid-search, Binary-search}}$
	\STATE \texttt{\small /* Select the input-output sequence of bins from Eq.~\ref{eq:seq-intervals}*/}
	\STATE $\Ie_{M+(b-1)n+1}\cdots, \Ie_{M+ b\cdot n} \leftarrow\textsc{SelectIntervals} (\Tst,\Tend)$
	\STATE $\Ecal([n]) \leftarrow \emptyset$
	\FOR{$j$ in $[n]$}
	    \STATE $C_{\min}, C_{\max} \leftarrow \textsc{ComputeConfidenceParams}(j)$
        \STATE \textsc{SetModelStates}$(\Ecal([j-1]))$ 
		\STATE \texttt{\small /* Solve the optimization problem in Eq.~\ref{eq:opt-model-g}*/}
	    \STATE $\Ecal(j)\leftarrow  \textsc{OptimizeInBin}(j,C_{\min},C_{\max}, \texttt{search}) $
%
	\ENDFOR
	\STATE $\Ecal' \leftarrow \textsc{EnumerateEvents}(\Ecal,\Tst,\Tend)$
	\STATE \text{Return} $\Ecal'$
	\end{algorithmic}
\end{algorithm}
\setlength{\textfloatsep}{0pt}
\begin{algorithm}[t]                    
	\caption{It implements \textsc{OptimizeInBin}}
	\label{alg:opt}
	\begin{algorithmic}[1]
	\STATE \textbf{Input: } The underlying bin $j$, $C_{\min},C_{\max}, \texttt{search}$.
	\STATE \textbf{Output: }  Predicted events $\{(\emk,\et)\}$ in the $j$-th bin.
 	\STATE $\Eb(C_{\min}: C_{\max}) \leftarrow \emptyset$
	\STATE $\Mb(C_{\min}: C_{\max}) \leftarrow \emptyset$
	\IF{\texttt{search} is \texttt{Grid-search}}
	\FOR{$C$ in $[C_{\min},C_{\max}]$}
		\STATE \texttt{\small /* Solve the optimization problem in Eq.~\ref{eq:opt-model-g}*/}
        \STATE ${\set{\ee _i}, \Mcal_{C}} \leftarrow \textsc{OptimizeForEvents}(C)$
        \STATE $\Eb(C)\leftarrow \set{\ee _i}$
        \STATE $\Mb(C)\leftarrow {\Mcal_{C'}}$
	\ENDFOR
   \STATE \texttt{\small /* Solve the optimization problem in Eq.~\ref{eq:Cstar}*/}
	\STATE $C^*\leftarrow \textsc{OptimizeForCounts}(\Mb, p_{\ephi})$ 
	\ENDIF
\IF{\texttt{search} is \texttt{Binary-search}}
	   \WHILE{low $<$ high}
        \STATE $\text{mid}_1 = \frac{\text{high}+\text{low}}{2}$, $\text{mid}_2=\text{mid}_1+1$.
        \STATE \texttt{\small /* Solve the optimization problem in Eq.~\ref{eq:opt-model-g}*/}
        \STATE $\cdots, \Mcal_{\text{mid}_1} \leftarrow \textsc{OptimizeForEvents}(\text{mid}_1)$
        \STATE \texttt{\small /* Compute $\Lcal_{\bullet}$ with Eq.~\ref{eq:lcalc}*/}
        \STATE $\Lcal_{\text{mid}_1}\leftarrow \textsc{ComputeLikelihood}(\Mcal_{\text{mid}_1}, \text{mid}_1)$
        \STATE \texttt{\small /* Solve the optimization problem in Eq.~\ref{eq:opt-model-g}*/}
        \STATE $\cdots, \Mcal_{\text{mid}_2} \leftarrow \textsc{OptimizeForEvents}(\text{mid}_2)$
        \STATE \texttt{\small /* Compute $\Lcal_{\bullet}$ with Eq.~\ref{eq:lcalc}*/}
        \STATE $\Lcal_{\text{mid}_2}\leftarrow \textsc{ComputeLikelihood}(\Mcal_{\text{mid}_2}, \text{mid}_2)$
        
        \IF{$\Lcal_{\text{mid}_1} > \Lcal_{\text{mid}_2}$} 
        \STATE high $=\text{mid}_1$
        \ELSE
        \STATE low $=\text{mid}_2$
        \ENDIF
        \ENDWHILE
        \STATE  $C^*=\text{high}$  \ \ \ \ \ \ \texttt{\small /* when high $=$ low */}
	\ENDIF
	\STATE \text{Return} $\Eb(C^*)$
	\end{algorithmic}
\end{algorithm}
We summarize this protocol in Algorithm~\ref{alg:basic-inference} and Algorithm~\ref{alg:opt} with $\texttt{search}=\texttt{Grid-search}$ (line no. 5--12 in Algorith~\ref{alg:opt}). Here, \textsc{SetModelStates}$(\Ecal([j-1]))$ sets the states $\hb_\bullet$ upto $j-1$ bins and then do a random run to draw  $C_{\max}$ dummy events from the $j$-th bin to setup 
to set the  $p_{\etheta}$.

\xhdr{Binary search based inference method}
The grid search inference method described above suffers from $O(C_{\max}-C_{\min})$ complexity per bin, since \textsc{OptimizeInBin}(..., , \texttt{Grid-search}) requires an exhaustive search over the set $[C_{\min}, C_{\max}]$. 
However, we design a more efficient search method by adopting two strategies. 
First, we start with an informed choice of
the confidence set \ie $[C_{\min}, C_{\max}]$. 
Second, instead of solving the optimization problem ~\eqref{eq:opt-model-g} for all candidates in the confidence set,
we use binary-search method to find out the optimal $C^*$. More specifically, we proceed as follows.
For the $i$-th bin $\Ie_{M+(b-1)n+i}$, we first solve the unconstrained
variant of the optimization problem~\eqref{eq:opt-model-g} to obtain $\Mcal_{C'}$, where $C'$
is the count of events within that bin. On the other hand, we independently maximize the likelihood of the count model \ie,
\begin{align}
 C'' =\argmax_{C} p_{\ephi}(  C_{{M+(b-1)n+i}}\hspace{-1mm}  = C| \Hcal_{(M+(b-1)n+1)\Delta} ).
\end{align}
Due to the unimodal convex nature of the count model $p_{\ephi}$, one can show that the optimal solution $C^*$ (Eq.~\ref{eq:Cstar})
lies between $C'$ and $C''$, which allows us to set the bounds of the confidence set with prior knowledge. To this end, we have
 $C_{\min} = \min\set{C',C''} $ and  $C_{\max} = \max\set{C',C''} $. Next, we observe that for a unimodal count distribution, we have
 \begin{align}
  \Lcal_{C_{\min}} < \Lcal_{C_{\min}+1} < \cdots < \Lcal_{C^*} > \Lcal_{C^*+1} >\cdots > \Lcal_{C_{\max}} 
 \end{align}
where, 
\begin{align}
 \Lcal_{C} =  \Mcal_{C} + p_{\ephi}(  C_{{M+(b-1)n+i}}\hspace{-1mm}  = C\given \Hcal_{(M+(b-1)n+1)\Delta} ). \label{eq:lcalc}
\end{align}
Therefore, we can select $C^*$ using a binary search algorithm as given in Algorithm~\ref{alg:opt} (line no. 15--36).
\fi
\end{document}

%% file: per_bin_results.tex
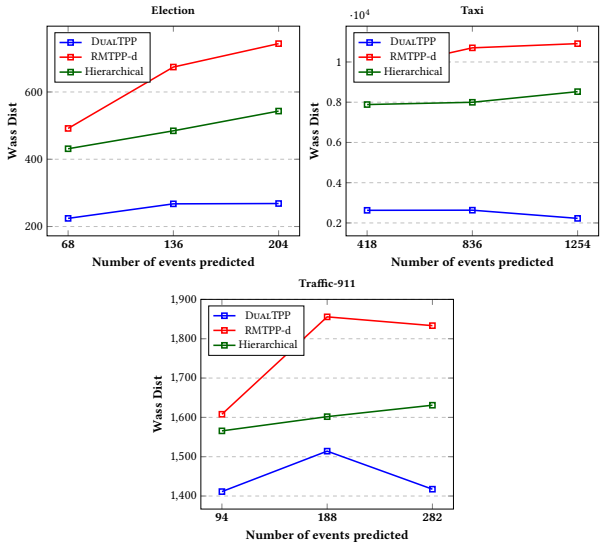
\begin{figure}
\begin{tikzpicture}[scale=0.49]
\begin{axis}[
    title={\textbf{Election}},
    xlabel={\textbf{Number of events predicted}},
    ylabel={\textbf{Wass Dist}},
    legend pos=north west,
    ymajorgrids=true,
    grid style=dashed,
    xticklabels={\textbf{68},\textbf{136},\textbf{204}},
    xtick={1,2,3},
    label style={font=\large},
    tick label style={font=\large}
]
	\addplot[
	color=blue,
	mark=square,
	very thick
	]
	coordinates {
	(1, 224.223)(2, 267.415)(3, 268.352)
	};
	\addplot[
	color = red,
	mark = square,
	very thick
	]
	coordinates {
	(1,491.608)(2, 674.101)(3,743.715)
	};
	\addplot[
	color= green!40!black!100,
	mark= square,
	very thick
	]
	coordinates {
	(1, 431.225)(2, 484.511)(3, 543.219)
	};
    \legend{\RmtppMseOpt, RMTPP-d, \RmtppMseOptComp}
\end{axis}
\end{tikzpicture}
\begin{tikzpicture}[scale=0.49]
\begin{axis}[
    title={\textbf{Taxi}},
    xlabel={\textbf{Number of events predicted}},
    ylabel={\textbf{Wass Dist}},
    legend pos=north west,
    ymajorgrids=true,
    grid style=dashed,
    xticklabels={\textbf{418},\textbf{836},\textbf{1254}},
    xtick={1,2,3},
    label style={font=\large},
    tick label style={font=\large}
]
	\addplot[
	color=blue,
	mark=square,
	very thick
	]
	coordinates {
	(1, 2628.136)(2, 2634.496)(3, 2227.464)
	};
	\addplot[
	color = red,
	mark = square,
	very thick
	]
	coordinates {
	(1,9541.025)(2, 10701.97)(3,10911.186)
	};
	\addplot[
	color= green!40!black!100,
	mark= square,
	very thick
	]
	coordinates {
	(1, 7883.996)(2, 7996.817)(3, 8525.384)
	};
    \legend{\RmtppMsevarOpt, RMTPP-d, \RmtppMsevarOptComp}
\end{axis}
\end{tikzpicture}


\begin{tikzpicture}[scale=0.49]
\begin{axis}[
    title={\textbf{Traffic-911}},
    xlabel={\textbf{Number of events predicted}},
    ylabel={\textbf{Wass Dist}},
    legend pos=north west,
    ymajorgrids=true,
    grid style=dashed,
    xticklabels={\textbf{94},\textbf{188},\textbf{282}},
    xtick={1,2,3},
    label style={font=\large},
    tick label style={font=\large}
]
	\addplot[
	color=blue,
	mark=square,
	very thick
	]
	coordinates {
	(1, 1411.255)(2, 1514.128)(3, 1417.38)
	};
	\addplot[
	color = red,
	mark = square,
	very thick
	]
	coordinates {
	(1,1607.956)(2, 1855.662)(3, 1833.31)
	};
	\addplot[
	color= green!40!black!100,
	mark= square,
	very thick
	]
	coordinates {
	(1, 1565.682)(2, 1601.843)(3, 1630.937)
	};
    \legend{\RmtppMseOpt, RMTPP-d, \RmtppMseOptComp}
\end{axis}
\end{tikzpicture}

\caption{Long term forecasting of \our, RMTPP-d and \RmtppMsevarOptComp across three datasets in terms of WassDist. RMTPP-d is just RMTPP with Gaussian density instead of intensity. X-axis denotes the average number of events in the gold since the known history $T$ and Y-axis denotes the Wasserstein distance between gold and predicted events.}
\label{fig:forecasting}
\end{figure}

%% file: sensitivity.tex
\begin{figure}[]
\begin{tikzpicture}[scale=0.49]
\begin{axis}[
    xlabel={\textbf{Bin Size ($\Delta$)}},
    ylabel={\textbf{Wass Dist}},
    legend pos=north west,
    ymajorgrids=true,
    grid style=dashed,
    xticklabels={\textbf{15 mins}, \textbf{30 mins}, \textbf{1 hr}, \textbf{3 hrs}, \textbf{6 hrs}},
    xtick={1,2,3,4,5},
    label style={font=\large},
    tick label style={font=\large} 
]
	\addplot[
	color=blue,
	mark = square,
	very thick
	]
	coordinates {
	(1, 2410) (2, 2039) (3, 1923) (4, 3959) (5, 4994)
	};
	\addplot[
	color = red,
	mark = square,
	very thick
	]
	coordinates {
	(1, 3875) (2, 4466) (3, 5679) (4, 8823) (5, 9662)
	};
	\addplot[
	color= green!40!black!100,
	mark= square,
	very thick
	]
	coordinates {
	(1, 8310) (2, 6545) (3, 8838) (4, 11673) (5, 8765)
	};
     \legend{\RmtppMseOpt, RMTPP-d, \RmtppMseOptComp}
\end{axis}
\end{tikzpicture}

\caption{Long term forecasting comparison on Taxi dataset: X-axis denotes the bin size used to train the count model $p_{\phi}$ and Y-axis denotes the Wasserstein Distance between true and predicted events. }
\label{fig:sensitivity}
\end{figure}
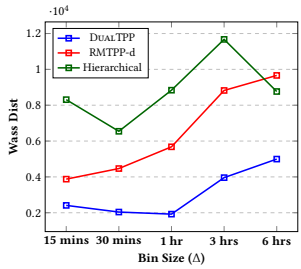
